\documentclass[twocolumn]{cinc}
\usepackage{amsmath}
\usepackage{multirow}
\usepackage{colortbl}
\usepackage{comment}
\usepackage{balance}
\usepackage{setspace} 
\usepackage{subcaption}
\usepackage{hyperref}
\usepackage{newtxmath}
\hypersetup{
    colorlinks = false,
    linkbordercolor = {1 0 0}}
\usepackage{url}

\usepackage{scalerel}
\usepackage{tikz}
\usetikzlibrary{svg.path}
\definecolor{orcidlogocol}{HTML}{A6CE39}
\tikzset{
  orcidlogo/.pic={
    \fill[orcidlogocol] svg{M256,128c0,70.7-57.3,128-128,128C57.3,256,0,198.7,0,128C0,57.3,57.3,0,128,0C198.7,0,256,57.3,256,128z};
    \fill[white] svg{M86.3,186.2H70.9V79.1h15.4v48.4V186.2z}
                 svg{M108.9,79.1h41.6c39.6,0,57,28.3,57,53.6c0,27.5-21.5,53.6-56.8,53.6h-41.8V79.1z M124.3,172.4h24.5c34.9,0,42.9-26.5,42.9-39.7c0-21.5-13.7-39.7-43.7-39.7h-23.7V172.4z}
                 svg{M88.7,56.8c0,5.5-4.5,10.1-10.1,10.1c-5.6,0-10.1-4.6-10.1-10.1c0-5.6,4.5-10.1,10.1-10.1C84.2,46.7,88.7,51.3,88.7,56.8z};
  }
}

\newcommand\orcidicon[1]{\href{https://orcid.org/#1}{\mbox{\scalerel*{
\begin{tikzpicture}[yscale=-1,transform shape]
\pic{orcidlogo};
\end{tikzpicture}
}{|}}}}

\begin{document}
\bibliographystyle{cinc}

\title{Zero-Shot Heart Rate Variability Forecasting from Consumer Wearables Using Time Series Foundation Models}

\author {Luukas Per{\"a}kyl{\"a}$^{\dagger}$\orcidicon{0009-0004-2443-4089}, Fahad Sohrab$^\ast$$^{\dagger}$\orcidicon{0000-0002-8080-4011}, Ville Hautam{\"a}ki$^\ast$\orcidicon{0000-0002-5885-0003}, Merja Hein{\"a}niemi$^\ast$\orcidicon{0000-0001-6190-3439}, Sui Huang$^\ast$\orcidicon{0000-0002-3545-4665}, Pekka Abrahamsson\orcidicon{0000-0002-4360-2226}$^{\dagger}$
\\
\ \\ %
$^\dagger$Tampere University, Tampere, Finland \\
$^\ast$University of Eastern Finland, Finland}

\maketitle

\begin{abstract}
Short-term Heart Rate Variability (HRV) forecasting could provide clinicians with actionable lead time for detecting autonomic dysfunction and adverse cardiac events. Consumer wearable devices generate fragmented, artifact-rich HRV signals that challenge conventional forecasting approaches. In this study, we evaluated the forecasting ability of three Time Series Foundation Models (TSFMs), TimesFM, Chronos, and MOIRAI, against traditional baselines (Mean, Exponential Smoothing, and Exponentially Weighted Moving Average) on real-world wearable data collected from 49 healthy individuals. To address data fragmentation, we introduce a variability-preserving imputation method that augments linear interpolation with locally adaptive stochastic noise, retaining physiological dynamics essential for accurate forecasting. The results show that TSFMs outperformed all baselines without fine-tuning, achieving average Mean Absolute Scaled Error (MASE) between 0.81 and 0.87 across TSFMs and both context lengths (32 and 64 time steps), with Chronos and TimesFM as the top models, though MOIRAI showed limited gains over baselines. With up to a 2-hour forecast horizon, the results establish a baseline for TSFMs' performance on a real-world dataset, highlighting domain-specific fine-tuning as a promising direction for clinical deployment.
\end{abstract}

\section{Introduction}
\begin{figure*}[t]
    \centering
    \includegraphics[width=0.7\linewidth]{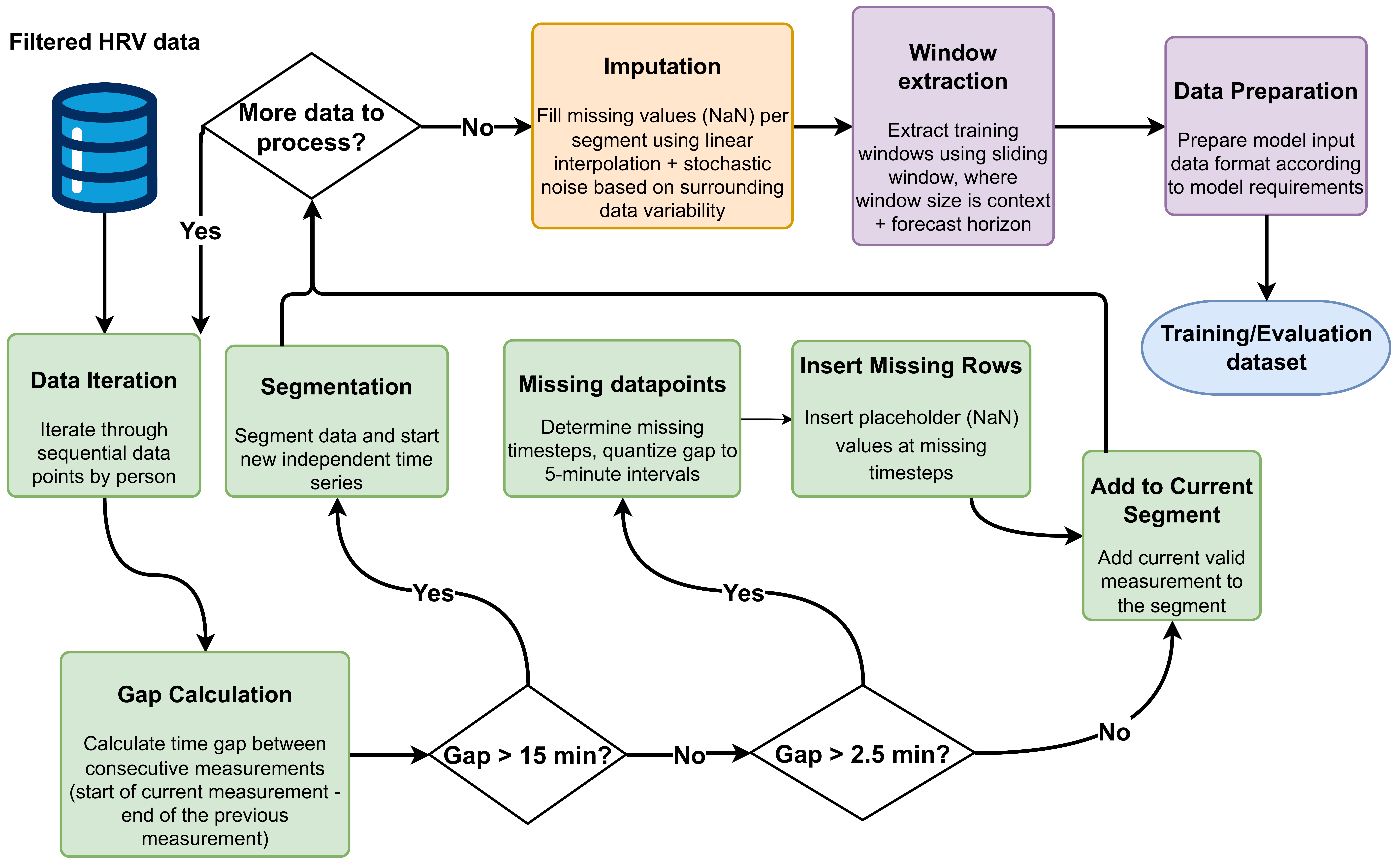}
    \caption{Data preprocessing pipeline for HRV time series, showing segmentation, gap handling,
    and imputation steps used to generate the final dataset for modelling.}
    \label{fig:hrv_pipeline}
\end{figure*}
Health care wearables, from smartwatches to biosensors, enable continuous and real-time physiological monitoring, tracking heart rate, sleep, and stress among other metrics, supporting personalized treatments, and can be used for the detection of diseases, such as atrial fibrillation \cite{huhn_impact_2022}. A promising direction within this space is the tracking of Heart Rate Variability (HRV) as a non-invasive tool to improve personal health outcomes \cite{li2023heart}. HRV reflects autonomic nervous system function and cardiovascular health, with higher HRV associated with better physiological adaptability and lower HRV indicating elevated risk \cite{shaffer2017overview, taskforce1996heart}. Accurate short-term HRV forecasting could therefore enable early alerting of autonomic dysfunction and adverse cardiac conditions, providing clinicians and monitoring systems with additional time for intervention~\cite{shaffer2017overview}.

The clinical value of HRV as a predictive signal is well established. Classifiers evaluating extracted HRV features have been shown to anticipate acute clinical events prior to symptom onset, such as paroxysmal atrial fibrillation \cite{maier_screening_2001} and ventricular fibrillation \cite{farhadi_sedehi_prediction_2017}. Using non-linear HRV features such as multiscale entropy, machine learning classifiers have successfully distinguished asymptomatic patients in the preclinical stages of cardiac autonomic neuropathy from healthy individuals \cite{cornforth_automated_2014}. Previous work has also validated deep learning on wearable data for detecting a broader range of medical conditions, including diabetes, high cholesterol, high blood pressure, and sleep apnea \cite{Ballinger_2018}.

However, these studies largely rely on high-fidelity, continuous clinical electrocardiogram (ECG) recordings rather than consumer-grade wearable signals. Translating HRV-based prediction to remote monitoring settings introduces significant challenges: data is inherently irregular and noisy due to device charging, motion artifacts, and non-wear periods, while limited continuous recording windows constrain the available context. Robust wearable-derived HRV can successfully predict major adverse cardiovascular events \cite{orini_wearable_2023}, and continuous photoplethysmography (PPG) measurements allow for real-time ischemic risk assessment, though mitigating signal noise remains a hurdle for these task-specific models \cite{tasmurzayev_wearable_2025}.

To overcome the limitations of task-specific, single-modality approaches, generalized Time Series Forecasting Models (TSFMs) provide an architecture explicitly designed to process heterogeneous, multivariate physiological signals generated by modern consumer devices \cite{luo_toward_2024}. Models such as TimesFM \cite{das2024timesfm}, Chronos \cite{ansari2024chronos}, and MOIRAI \cite{woo2024moirai} are designed for robust zero-shot generalization across diverse domains without task-specific fine-tuning, making them natural candidates for wearable HRV forecasting. This study investigates whether these models can accurately forecast wearable HRV without fine-tuning, how context and horizon length affect forecasting accuracy, and whether they outperform classical statistical baselines on real-world data.

\section{Methodology}
This section describes the dataset, preprocessing pipeline, and experimental setup used to evaluate the forecasting models. We first outline the characteristics of the wearable HRV data and the imputation strategy used to handle missingness, followed by the models and evaluation protocol.
\subsection{Data and Imputation}
We evaluated models on a public real-world HRV dataset comprising wearable-based measurements from 49 healthy individuals collected over four weeks using Samsung Galaxy Watch Active 2 devices
\cite{baigutanova2025continuous}. The device uses Photoplethysmography (PPG), a common but
noise-susceptible method for digital biomarker measurement \cite{li2023heart}. Among HRV
time-domain metrics, the Root Mean Square of Successive Differences (RMSSD) robustly captures
short-term parasympathetic activity and is particularly suited for short, PPG-derived wearable
recordings \cite{shaffer2017overview, taskforce1996heart}. Only pre-filtered (missingness score
$\leq 0.35$) RMSSD data was used, computed from 5-minute PPG segments.

The dataset reflects the highly fragmented nature of real-world consumer wearable data. After applying a 15-minute gap threshold, participants contributed a mean of 109 continuous segments each, with longest continuous recordings averaging 4.4 hours and a maximum of approximately 14.75 hours, driven by nightly charging cycles, motion artifacts, and non-wear periods \cite{baigutanova2025continuous}. The chosen context lengths of 32 and 64 steps (2.7 and 5.3 hours) reflect these real-world recording limitations rather than arbitrary design choices.

To generate regularly sampled time series suitable for the models, smaller gaps were imputed. Standard imputation methods (linear interpolation, KNN, cubic splines, and forward-fill) can produce unrealistically smooth sequences that affect both metrics and model learning. To address this, we augmented linear interpolation with locally adaptive stochastic noise:

\begin{equation}
\hat{x}_t = x_t^{\text{linear}} + \epsilon_t.
\end{equation}
Here $\hat{x}_t$ is the imputed value at time $t$, $x_t^{\text{linear}}$ is the linearly
interpolated value, and $\epsilon_t$ is zero-mean Gaussian noise with standard deviation
$\sigma_{\text{local}}(1 + |\beta|)$. Local standard deviation $\sigma_{\text{local}}$ is
computed from a 5-step window, $\beta$ represents the local slope, and noise is clipped at
$3\sigma_{\text{local}}$ to prevent outliers. The multiplicative term $(1 + |\beta|)$ adaptively
modulates noise variance based on local slope magnitude, preserving physiological variability
structure essential for accurate forecasting. The preprocessing pipeline is illustrated in
Figure~\ref{fig:hrv_pipeline}.
 
 \begin{figure*}[htbp]
    \centering
    \includegraphics[width=\linewidth]{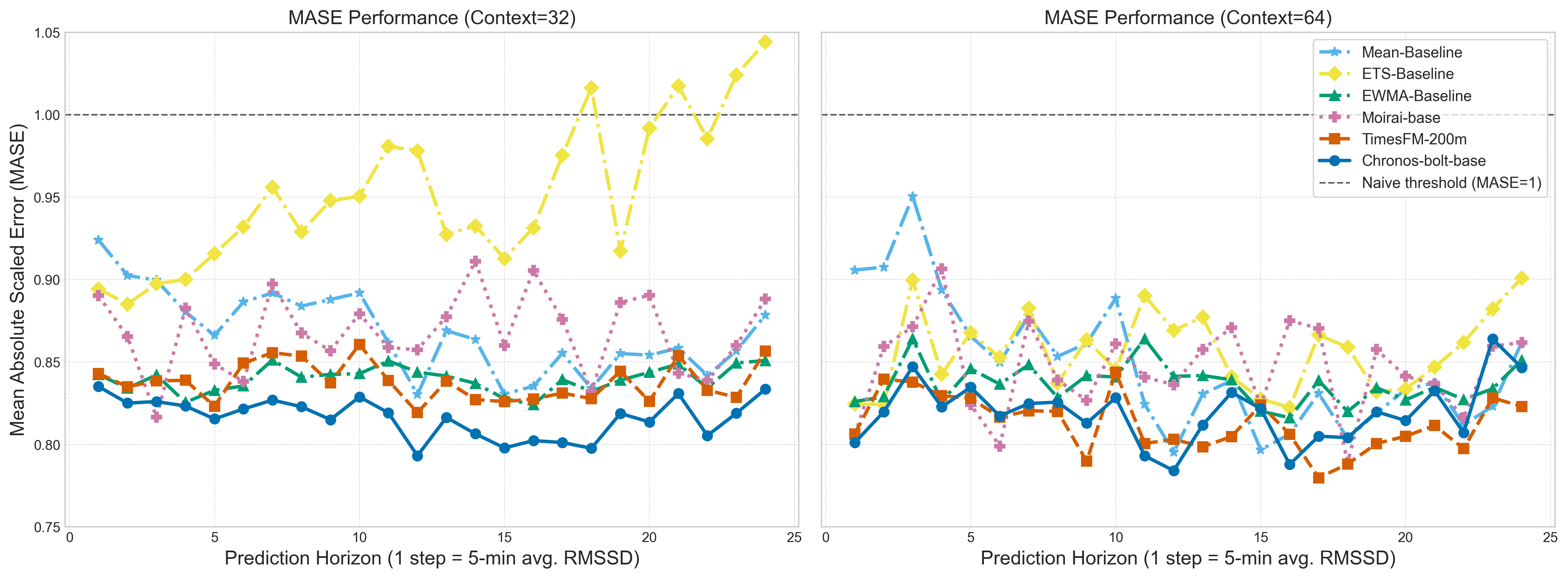}
    \caption{MASE across forecast horizons for context length 32 (left) and 64 (right). 
    The dashed line denotes the naive forecasting threshold (MASE\,${=}$\,1.0). 
    ETS exceeds the plot range at longer horizons in context 32.}
    \label{fig:mase_horizons}
\end{figure*}
 
\subsection{Models and Experimental Setup}
 
Three foundation models were evaluated without fine-tuning: TimesFM-1.0-200m
\cite{das2024timesfm}, Chronos-bolt-base \cite{ansari2024chronos}, and 
MOIRAI-1.0-R-base \cite{woo2024moirai}. These were benchmarked against traditional baselines:
Mean, Exponentially Weighted Moving Average (EWMA, $\alpha{=}0.3$) \cite{holt2004forecasting},
and Exponential Smoothing (ETS) with additive damped trend. Two context lengths were tested: 32
steps (2.7 hours) and 64 steps (5.3 hours), each step representing a 5-minute RMSSD average.
Forecast horizons ranged from 1 to 24 steps (up to 2 hours ahead). Probabilistic model outputs
(Chronos, MOIRAI) were reduced to the median for point-forecast comparability with TimesFM.
Evaluation metrics were Mean Absolute Error (MAE), Root Mean Square Error (RMSE) \cite{holt2004forecasting}, and Mean Absolute Scaled Error (MASE) \cite{Hyndman2006}.

As this is a zero-shot evaluation, no training or fine-tuning was performed on any subset of this dataset; the entire dataset serves as the evaluation set for all foundation models. The classical baselines require no training: EWMA uses a fixed smoothing factor of $\alpha=0.3$ chosen a priori. ETS smoothing parameters are fitted independently on each context window with no shared training across subjects, ensuring no cross-subject data leakage.

\section{Results}
 
Tables~\ref{table:32_avg_table} and \ref{table:64_avg_table} summarise average forecast
performance across all horizons. Chronos-bolt-base achieves the lowest MAE, RMSE, and MASE at
context length 32, while TimesFM-200m achieves the best MAE and MASE at context length 64.
Increasing context length from 32 to 64 improves performance for all foundation models, with
TimesFM gaining the largest improvement and surpassing Chronos at the longer context.
Among traditional baselines, EWMA remains competitive at short horizons but is consistently
outperformed by foundation models for longer predictions. MOIRAI did not significantly outperform the Mean baseline at either context length (p=0.953 and p=0.794 respectively), which may reflect sensitivity to the short, highly fragmented segment lengths characteristic of consumer wearable data, where its patch-based architecture may not have sufficient continuous context to operate effectively.
 
\begin{table}[h!]
\centering
\small
\caption{Average performance for context length 32, over all forecast horizons (1--24).}
\label{table:32_avg_table}
\begin{tabular}{|l|r|r|r|}
\hline
Model & MAE & RMSE & MASE \\
\hline
Mean           & 16.74 & \underline{26.04} & 0.87 \\
\hline
EWMA           & 16.19 & 26.18 & \underline{0.84} \\
\hline
ETS            & 18.37 & 34.80 & 0.95 \\
\hline
MOIRAI         & 16.73 & 26.68 & 0.87 \\
\hline
TimesFM        & \underline{16.16} & 26.55 & \underline{0.84} \\
\hline
Chronos        & \textbf{15.74} & \textbf{25.64} & \textbf{0.82} \\
\hline
\end{tabular}
\end{table}
 
\begin{table}[h!]
\centering
\small
\caption{Average performance for context length 64, over all forecast horizons (1--24).}
\label{table:64_avg_table}
\begin{tabular}{|l|r|r|r|}
\hline
Model & MAE & RMSE & MASE \\
\hline
Mean           & 15.99 & \textbf{25.80} & 0.85 \\
\hline
EWMA           & 15.77 & 26.34 & 0.84 \\
\hline
ETS            & 16.15 & 27.88 & 0.86 \\
\hline
MOIRAI         & 15.94 & 26.34 & 0.85 \\
\hline
TimesFM        & \textbf{15.32} & 26.41 & \textbf{0.81} \\
\hline
Chronos        & \underline{15.44} & \underline{26.24} & \underline{0.82} \\
\hline
\end{tabular}
\end{table}
 
Figure~\ref{fig:mase_horizons} illustrates performance trends across all horizons, confirming that
foundation models maintain their advantage throughout and remain below the MASE 1.0 threshold
relative to naive forecasting. Forecasting accuracy decreased only modestly with longer horizons,
likely because the maximum horizon of 24 steps remains short relative to HRV's chaotic dynamics,
and because averaging error metrics over more forecast points reduces the impact of individual
outliers.

Paired t-tests (Table~\ref{tab:significance}) confirm that differences between foundation models
and traditional baselines are statistically significant for most comparisons. Chronos outperforms
EWMA across both contexts ($p{<}0.001$). Differences between TimesFM and Chronos are significant
only at context 32, with the two models becoming comparable at context 64 as TimesFM leverages
the longer history more effectively. 
 
\begin{table}[h!]
\centering
\caption{Statistical significance of key model comparisons (paired t-test, $n{=}24$ horizons,
$\alpha{=}0.05$). $\Delta$MAE is used as the test statistic; negative values indicate Model 1 is better.}
\label{tab:significance}
\scriptsize
\resizebox{\linewidth}{!}{%
\begin{tabular}{|l|l|r|r|r|r|}
\hline
\textbf{Model 1} & \textbf{Model 2} &
\multicolumn{2}{c|}{\textbf{Context 32}} &
\multicolumn{2}{c|}{\textbf{Context 64}} \\
\cline{3-6}
 & & \textbf{$\Delta$MAE} & \textbf{p-val} & \textbf{$\Delta$MAE} & \textbf{p-val} \\
\hline
TimesFM  & Chronos & 0.42  & $<0.001^*$ & -0.12 & 0.057 \\
TimesFM  & EWMA    & -0.03 & 0.450      & -0.46 & $<0.001^*$ \\
TimesFM  & Mean    & -0.58 & $<0.001^*$ & -0.67 & $<0.001^*$ \\
Chronos  & EWMA    & -0.45 & $<0.001^*$ & -0.33 & $<0.001^*$ \\
Chronos  & Mean    & -0.99 & $<0.001^*$ & -0.55 & $0.001^*$ \\
MOIRAI   & EWMA    &  0.54 & $<0.001^*$ &  0.17 & 0.121 \\
MOIRAI   & Mean    & -0.01 & 0.953      & -0.04 & 0.794 \\
EWMA     & Mean    & -0.54 & $<0.001^*$ & -0.21 & 0.141 \\
\hline
\end{tabular}}
\end{table}
\vspace{-5mm}
\section{Conclusions}
\vspace{-1mm}
HRV is a valuable non-invasive indicator of autonomic function and cardiovascular health,
increasingly accessible through consumer wearable devices. In this study, we evaluated the
zero-shot forecasting capability of TSFMs on real-world wearable HRV
data, benchmarking them against traditional statistical models. Our results demonstrate that foundation models outperform traditional baselines, with Chronos-bolt-base excelling at shorter contexts and TimesFM-200m at longer contexts. Importantly, foundation models maintain superior accuracy at longer forecast horizons, enabling earlier prediction of adverse autonomic events and providing greater lead time for clinical intervention. These results suggest Chronos and TimesFM are viable candidates for deployment in consumer cardiac monitoring pipelines with domain-specific fine-tuning, while MOIRAI's sensitivity to data fragmentation warrants further investigation. A key limitation is that the evaluation was restricted to healthy individuals. The primary remaining step is validation on patient cohorts with known cardiac conditions, which this study's baseline performance on healthy individuals is designed to support.  
 
\begin{spacing}{0.86}
\bibliography{refs}
\end{spacing}
 \vspace{-3mm}
\begin{correspondence}
Fahad Sohrab (fahad.sohrab@tuni.fi)\\
P.O. Box $553$, FI$-33014$, Tampere Finland\\
\end{correspondence}
\end{document}